%% file: acl_latex.tex
\pdfoutput=1

\documentclass[11pt]{article}

\usepackage[]{acl}

\usepackage{times}
\usepackage{latexsym}

\usepackage[T1]{fontenc}

\usepackage[utf8]{inputenc}

\usepackage{microtype}

\usepackage{array,multirow}
\usepackage{graphicx}
\usepackage{bm}
\usepackage{amsmath}
\usepackage{amssymb}
\usepackage{amsfonts}
\usepackage{url}
\usepackage{algorithm}
\usepackage[noend]{algpseudocode}
\usepackage {arydshln}

\renewcommand{\u}{\underline}

\title{
    Downstream Task-Oriented Neural Tokenizer Optimization \\ with Vocabulary Restriction as Post Processing    
}

\author{
    Tatsuya Hiraoka \quad\quad Tomoya Iwakura \\
    Fujitsu Limited \\ 
    \texttt{\{hiraoka.tatsuya, iwakura.tomoya\}@fujitsu.com}
}

\begin{document}
\maketitle
\begin{abstract}
This paper proposes a method to optimize tokenization for the performance improvement of already trained downstream models.
Our method generates tokenization results attaining lower loss values of a given downstream model on the training data for restricting vocabularies and trains a tokenizer reproducing the tokenization results.
Therefore, our method can be applied to variety of tokenization methods, while existing work cannot due to the simultaneous learning of the tokenizer and the downstream model.
This paper proposes an example of the BiLSTM-based tokenizer with vocabulary restriction, which can capture wider contextual information for the tokenization process than non-neural-based tokenization methods used in existing work.
Experimental results on text classification in Japanese, Chinese, and English text classification tasks show that the proposed method improves performance compared to the existing methods for tokenization optimization.
\end{abstract}

\input{sections/1_introduction}
\input{sections/2_related}
\input{sections/3_aspostproc}
\input{sections/4_proposed}
\input{sections/5_experiment}
\input{sections/6_discussion}
\input{sections/7_conclusion}

\section*{Acknowledgement}
This work was supported by JST, ACT-X Grant Number JPMJAX21AM, Japan.


\bibliography{anthology,custom}
\bibliographystyle{acl_natbib}


\end{document}

%% file: sections/1_introduction.tex
\section{Introduction}

Tokenization is an important pre-processing that is commonly used in various NLP tasks.
We can acquire the performance improvement of downstream tasks (e.g., text classification and machine translation) by selecting appropriate tokenization for each task~\cite{hiraoka2019stochastic,bostrom2020byte,shin2020biomegatron}.
Recent research introduced methods to seek appropriate tokenization for already trained downstream models~\cite{hiraoka2020optimizing, hiraoka2021joint}.
These methods attempt to find tokenization that improves the performance in a downstream task while the parameters of the downstream model are frozen after the training of the model.

OpTok~\cite{hiraoka2021joint} is a recent method to optimize tokenization for the trained downstream models.
Although OpTok can be applied for various downstream tasks, it has poor scalability because it cannot be used for the downstream models that do not use a unigram-based tokenizer~\cite{kudo2018sentencepiece}.
This is because OpTok is implemented for the simultaneous optimization of the downstream model and the unigram-based tokenizer.
If we can use more capable tokenization methods (e.g., the one using LSTM~\cite{hochreiter1997long}) other than the unigram-based ones, we might acquire more improvement by optimizing tokenization.

In this study, we propose a training strategy to use various tokenization methods for improving the accuracy of the already trained downstream model.
Our strategy dissolves the dependency between the downstream model and the tokenizer.
In concrete, we introduce a training strategy with two steps: (1) we create a training dataset composed of appropriate tokenizations (\S \ref{sec:creating_dataset}), and (2) we train a new tokenizer to reproduce the dataset (\S \ref{sec:train_tokenizer}).

In our framework, we can use various tokenization methods not limited to the unigram-based method.
For example, a representative tokenizer using a BiLSTM architecture~\cite{ma-etal-2018-state} can be incorporated into the proposed method.
Besides, we introduce a more effective BiLSTM-based tokenizer with vocabulary restriction that does not yield unknown tokens for the downstream model (\S \ref{sec:tokenizer_with_restriction}).

We conducted experiments on text classification in Chinese, Japanese, and English (\S \ref{sec:experiments}).
Besides, we use three tokenization methods for the downstream models: the unigram-based method~\cite{kudo2018sentencepiece}, the byte pair encoding-based method (BPE)~\cite{sennrich2016neural}, and the maximum matching-based method~\cite{song2020linear}.
Experimental results demonstrated that the proposed method overcomes the existing methods in many settings.
Also, the proposed method can reproduce the appropriate tokenization more accurately (\S  \ref{sec:reproducibility}) and does not yields unknown tokens other than unknown characters (\S \ref{sec:unk_rate}).

\begin{figure*}[ht]
\centering
\includegraphics[width=15.5cm]{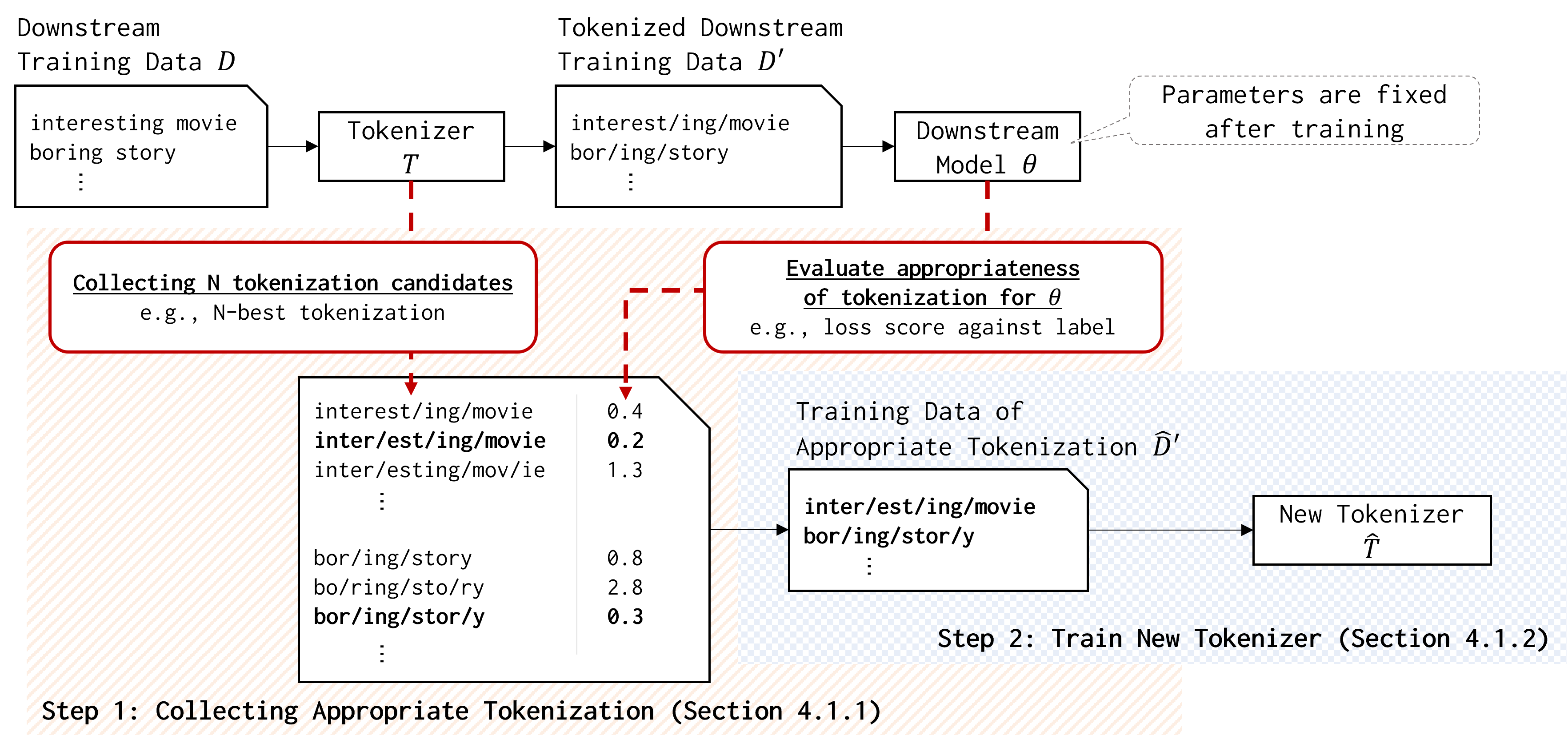}
\caption{
    Outline of the proposed training strategy with two steps: collecting appropriate tokenization (\S \ref{sec:creating_dataset}) and training the new tokenizer (\S \ref{sec:train_tokenizer}).
}
\label{fgr:outline}
\end{figure*}

%% file: sections/2_related.tex
\section{Related Work}
This study is in line with research about tokenization optimization.
\newcite{hiraoka2020optimizing} introduced a simultaneous optimization of a tokenizer and a downstream model, following research seeking appropriate tokenization for machine translation tasks~\cite{xu2008bayesian,chung2009unsupervised,salesky2020optimizing,xuanli2020dynamic}.
\newcite{hiraoka2021joint} modified this work for efficient learning.
Our study is different from this literature in focusing on more specific situations; instead of simultaneous learning of tokenization and downstream models, our work focuses on optimizing tokenization as \textbf{post-processing} where the downstream models are already trained and their parameters are no longer updated. 
Although \newcite{hiraoka2021joint} provides a small investigation about this setting, we dive into this task more deeply with the method specialized to the optimizing tokenization as post-processing and much more detailed examination for the first time.
This paper is also different from the existing work in the applicable tokenizer types for optimization.
The proposed method enables us to use various tokenizers including neural networks-based ones while existing work is limited to unigram-based tokenizers~\cite{kudo2018sentencepiece}.


%% file: sections/3_aspostproc.tex
\section{Task Definition}
This paper focused on a task to optimize tokenization for already trained downstream models as post processing for improving accuracy.
In this task, we presuppose a training data $D$ of the downstream task (e.g., text classification), a tokenizer $T$, and a downstream model $\theta$ trained with a training data $D'$ whose samples are tokenized with $T$.
$s \in D$ is a text comprising the training data and $s' \in D'$ is a text tokenized with $T$ where $s' = T(s)$.
We can use various tokenization methods for $T$ such as the unigram-based method~\cite{kudo2018sentencepiece}, the BPE~\cite{sennrich2016neural}, and the maximum matching-based method~\cite{song2020linear}.

We denote $N$ types of the possible tokenization variation of $s$ as $s'_1 ... s'_N$.
Here, the maximum number of tokenization variations is $N=2^{|s|-1}$ where $|s|$ is the number of characters comprising $s$.
$N$, in practice, falls in a reasonable range because the downstream model has a limited number of tractable tokens as the vocabulary.

The purpose of the task, tokenization optimization as post-processing, is to find a tokenizer $\hat{T}$ that yields tokenization which improves the performance of the trained downstream model $\theta$ on the downstream task.


%% file: sections/4_proposed.tex
\section{Proposed Method}

\subsection{Training Strategy}
The purpose of this study is to establish a strategy for optimizing tokenization without modifying the downstream model architecture.
Instead of incorporating a tokenizer into the downstream model as the conventional methods~\cite{hiraoka2020optimizing,hiraoka2021joint} do,
our method trains another tokenizer to optimize tokenization.
Concretely, our training strategy is composed of two steps as showen in Figure \ref{fgr:outline}:
(1) collecting tokenization results that improves the performance on the training split (\S \ref{sec:creating_dataset}) and (2) training a new tokenizer that reproduces \textit{better} the collected tokenization (\S \ref{sec:train_tokenizer}).
Because this two-step strategy enable us to train new tokenizer independently from the downstream model, we can use various types of tokenizer architectures for optimizing tokenization in addition to the unigram-based tokenizers.

\subsubsection{Collecting Tokenization with Small Loss}
\label{sec:creating_dataset}
In the first step, we construct a new training data of tokenization $\hat{s}' \in \hat{D}'$ from $D$.
Here, $\hat{s}'$ is a tokenization with which the downstream model $\theta$ performs best among $N$ different tokenizations on the training data $D$.
We can collect such the tokenization $\hat{s}'$ by inputting $N$ various tokenization ${s'_1 ... s'_N}$ into the model $\theta$, calculating the loss value against the ground truth signal, and then, picking the tokenization with the lowest loss values as the following:
\begin{align}
    \hat{s}' = \mathop{argmin}_{1 \leq n \leq N} f(s'_n; \theta),
\end{align}
where $f(\cdot)$ is the loss function.
For obtaining $N$ different tokenizations, we can exploit $N$-best tokenization~\cite{nagata1994stochastic} or tokenization sampling with subword regularization~\cite{kudo2018subword,provilkov2019bpe,hiraoka2022maxmatch}.

\subsubsection{Train Tokenizer Reproducing $\hat{T}$}
\label{sec:train_tokenizer}
In the second step, we train a new tokenizer $\hat{T}$ with the newly created training data $\hat{D}'$.
$\hat{D}'$ is composed of tokenizations that lead to the higher performance (the lower loss values) for its downstream model $\theta$ in the training data.
Therefore, we expect that the new tokenizer $\hat{T}$ trained on $\hat{D}'$ could yield appropriate tokenizations even in validation or test splits.
Because the tokenizer $\hat{T}$ is newly trained on $\hat{D}'$, it is completely independent from the downstream model.
This training strategy enables us to use various tokenization methods for $\hat{T}$ such as more representative tokenizers with neural networks.
Using such the strong tokenizer for optimizing tokenzation might improves the performance of the downstream model as the post processing.

\begin{figure}[t]
\centering
\includegraphics[width=7.7cm]{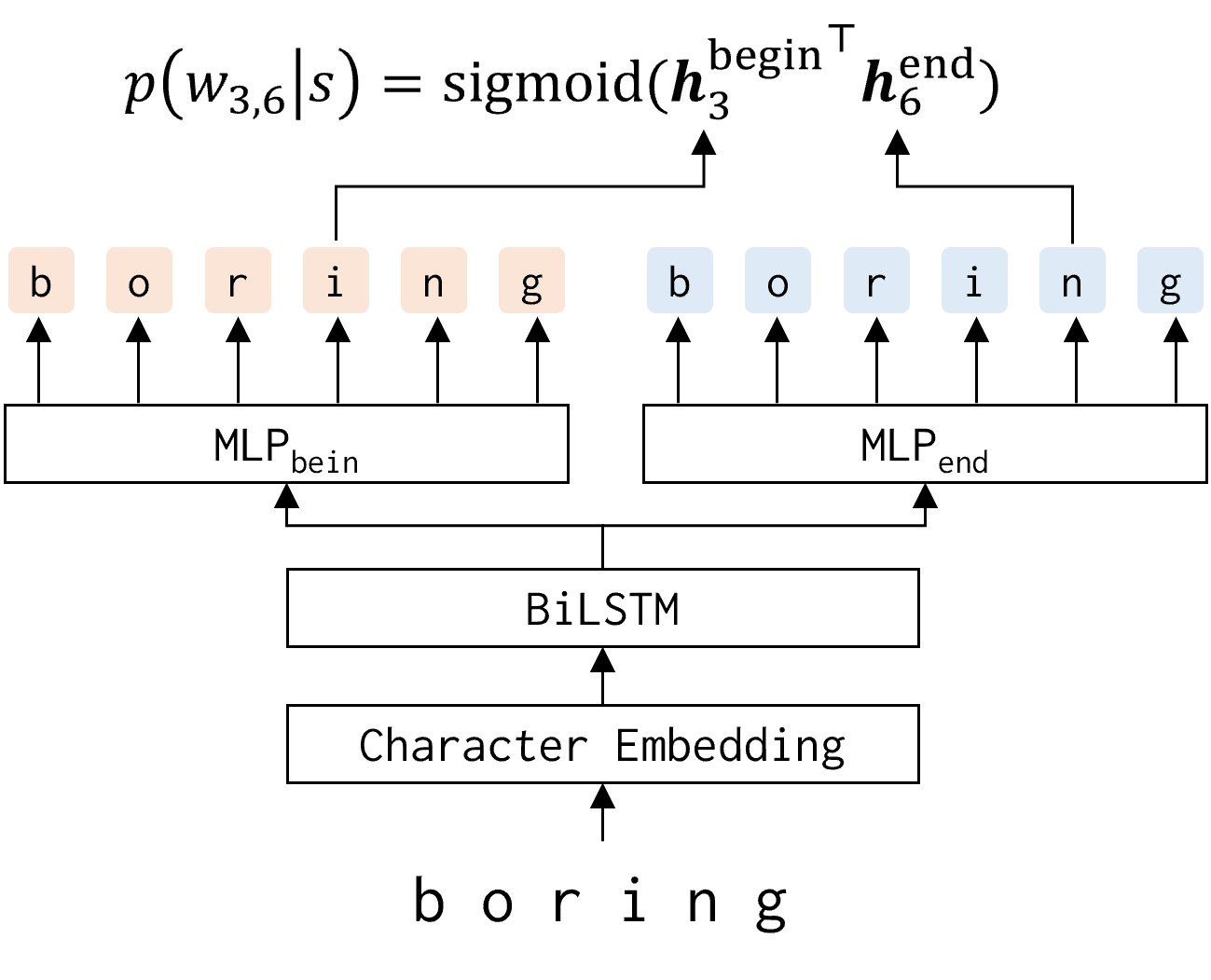}
\caption{
    Calculation outline of the probability of token "ring" in a text "boring" with the tokenizer introduced in \S \ref{sec:tokenizer_with_restriction}.
    This figure shows the case where "ring" is in a vocabulary.
}
\label{fig:tokenizer}
\end{figure}

\begin{figure}[t]
\centering
\includegraphics[width=7.7cm]{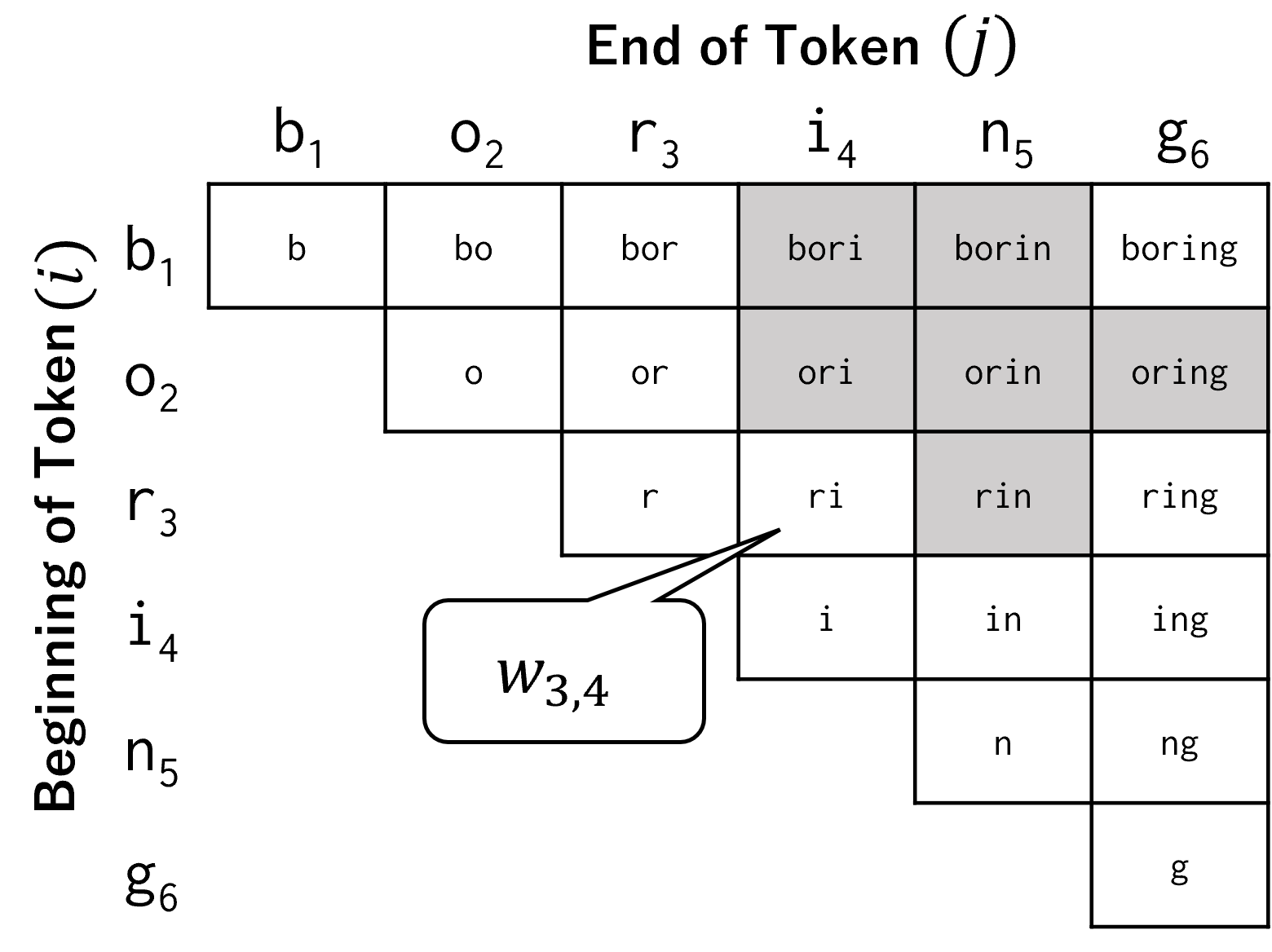}
\caption{
    Outline of vocabulary restriction used in our method (Eq. \eqref{eq:prob}) represented in a table style.
    Shaded cells represent unknown tokens that are not in the vocabulary.
}
\label{fig:table}
\end{figure}

\subsection{Vocabulary-Restricted Neural Tokenizer}
\label{sec:tokenizer_with_restriction}

We can use various types of architectures for the tokenizer $\hat{T}$ that reproduces tokenizations in $\hat{D}'$.
We expect that the larger performance improvement could be obtained with the more capable tokenization architecture for $\hat{T}$.

One popular choice of such a representative tokenizer is a neural network-based architecture with BI-tagging schema~\cite{chen2015long,young2018recent}.
With this architecture, the tokenization is expressed with B (representing the beginning of a word) and I (representing the intermediate of a word) tags.
And the neural network-based tokenizer is trained with these tags as a sequential labeling task.
Although these methods with BI-tag schema succeeded in many tokenization benchmarks, 
we cannot utilize them as it is because they output unknown tokens for the downstream model.
In other words, such representative tokenization architectures output tokens that are not in the vocabulary $V_\theta$ of the downstream model $\theta$.
These unknown tokens cannot be encoded into token embeddings appropriately, and this would damage the final performance of the downstream task.

To address this problem, we propose a solution with an architecture that can directly take $V_\theta$ into account inspired by \newcite{higashiyama2019incorporating}.
The proposed tokenizer has a hard restriction for the output so that it cannot yield tokenization including unknown tokens.
This method calculates the probability of a tokenization result $s'$ of a sentence $s$ as the following:
\begin{align}
    p(s') = \prod_{w \in s'}p(w|s),
\end{align}
where $w$ is a token in $s'$.
For a sentence with $K$ characters $s=c_1 ... c_K$, 
we calculate the probability of the token $w_{i, j}$ that begins with the $i$-th character and ends with $j$-th character as
\begin{align}
\mathbf{h}_k &= \mathrm{BiLSTM}(\mathbf{v}_{c_1} ... \mathbf{v}_{c_K})_k, \\
\mathbf{h}^{(\mathrm{begin})}_k &= \mathrm{MLP_{begin}}(\mathbf{h}_k), \\
\mathbf{h}^{(\mathrm{end})}_k &= \mathrm{MLP_{end}}(\mathbf{h}_k), \\
p(w_{i, j}|s) &= 
\begin{cases}
\sigma(
{\mathbf{h}^{(\mathrm{begin})}_i}^\top
\mathbf{h}^{(\mathrm{end})}_j
) \quad&\mathrm{if} w_{i, j} \in V_\theta \\
0 \quad&\mathrm{otherwise}
\end{cases}\label{eq:prob}.
\end{align}
Here, $\mathrm{BiLSTM(\cdot)_k}$ is an operation to obtain the outputs of BiLSTM~\cite{hochreiter1997long, graves2005framewise} corresponding to the $k$-th input $\mathbf{v}_{c_k}$.
$\mathbf{v}_{c_k}$ is a character embedding corresponding to $c_k$.
$\mathrm{MLP_{begin}}$ and $\mathrm{MLP_{end}}$ are distinct multilayer perceptrons (MLPs) to represent the beginning and end points of tokens, respectively.
$\sigma(\cdot)$ denotes the Sigmoid function.

We can calculate $p(w_{i, j}|s)$ at once for each $s$ because this operation can be represented as an upper triangular matrix shown in \ref{fig:table}.
As shown in the figure and Eq. \eqref{eq:prob}, the probabilities of tokens that are not included in the downstream vocabulary $V_\theta$ are masked with zero values.
This hardly restricts the output of the tokenizer so that it cannot yield tokenization with unknown tokens that are not included in $V_\theta$.

Given a training sample of tokenization $\hat{s}'$, the tokenizer is optimized to minimize the following loss value.
\begin{align}
\mathcal{L}_{\hat{s}'} = -{\sum_{w \in \hat{s}'}{\log p(w|s)}}.
\end{align}
In the inference step, we obtain the tokenization that maximizes $\sum_{w \in s'}\log(p(w|s))$ with the Viterbi algorithm~\cite{viterbi1967error}.

%% file: sections/5_experiment.tex
\section{Experimental Setttings}
\label{sec:experiments}

\input{tables/datasets}
\input{tables/downstream_result.tex}

\subsection{Dataset}
The proposed method is applicable to various downstream tasks if we can evaluate the effectiveness of tokenization candidates for the downstream models (e.g., loss values).
In this paper, we employed text classification as the downstream task following the existing work~\cite{hiraoka2020optimizing,hiraoka2021joint}.
Optimizing tokenization would have a larger effect on the performance, especially for languages without whitespaces indicating word boundaries.
In this experiment, we employed Chinese and Japanese text classification datasets.
In addition to these unsegmented languages, we also evaluate our method on the dataset in English that is accompanied by whitespaces indicating word boundary.
Table \ref{tbl:dataset} summarizes the statistics of each dataset.

For the Chinese datasets, \textbf{Weibo} is a sentiment classification dataset on Chinese short text SNS\footnote{\url{https://github.com/wansho/senti-weibo}}.
\textbf{Genre} is a genre classification of products from reviews and \textbf{Rate} is a rate classification of reviews.
Both Genre and Rate datasets are constructed from the same review corpus of the E-commerce service~\cite{zhang2015daily}.

For the Japanese datasets, \textbf{Twitter}~\cite{suzuki2019filtering} is a sentiment classification dataset on the Japanese Twitter corpus.
Although we use this dataset following the existing literature~\cite{hiraoka2020optimizing,hiraoka2021joint}, many tweet data referred to in this dataset have already been unavailable.
Considering the reproducibility of the experiments, we also employed an additional sentiment classification dataset on Twitter named \textbf{WRIME}~\cite{kajiwara2021wrime}, which is a reproducible dataset including tweet data.

In addition to the Asian language datasets, we used English \textbf{Twitter}\footnote{\url{https://www.kaggle.com/c/twitter-sentiment-analysis2}} that is a sentiment classification on English Twitter.

\subsection{Downstream Model}
\label{sec:downstream_model}
As the downstream model $\theta$ for text classification, we used a simple classification model that encodes an input text with BiLSTM and outputs a label-sized vector with MLP and the softmax function.
Before the encoding, the tokens in the input text are encoded into token embeddings whose size is 64.
The number of BiLSTM layers was 1, and the size of hidden vectors of BiLSTM was 256 for both forward and backward cells.
We trained the downstream model on the training data with 20 epochs and selected a model with the best performance on the validation data as $\theta$.

To demonstrate the applicability of our method to donwstream models trained with various tokenizers, we employed three popular tokenizers to train the downstream model $\theta$.
\textbf{Unigram} is a unigram-based tokenizer.
We utilized the unigram mode of SentencePiece~\cite{kudo2018sentencepiece}.
\textbf{BPE}~\cite{sennrich2016neural} is a merge-based tokenizer.
We used the BPE mode of SentencePiece.
\textbf{MaxMatch}~\cite{song-etal-2021-fast} is a maximum matching algorithm-based tokenizer.
We employed the implementation by HuggingFace known as BertTokenizer (WordPiece).

To obtain $N$ tokenization candidates for the proposed method, we used $N$-best tokenization~\cite{nagata1994stochastic} for Unigram.
Because there is no method to obtain $N$-best tokenization for BPE and MaxMatch, we collect $N$ candidates by sampling tokenization with BPE-Dropout~\cite{provilkov2019bpe} and MaxMatch-Dropout~\cite{hiraoka2022maxmatch}.

For all settings, we set the size of vocabulary $V_\theta$ as 16,000.
During the training of the downstream model, we applied subword regularization with $\alpha=0.2$ for \textbf{Unigram} and $p=0.1$ for \textbf{BPE} and \textbf{MaxMatch}\footnote{$\alpha$ is a hyperparameter for smoothness of sampling distribution~\cite{kudo2018subword} . $p$ is a dropout rate for BPE/MaxMatch-Dropout~\cite{provilkov2019bpe, hiraoka2022maxmatch}.}.

\subsection{Baselines of Optimized Tokenizers}
We compared the tokenizer introduced in \S \ref{sec:tokenizer_with_restriction} with three baselines.

\paragraph{$\text{Unigram}^\text{OPT}$} We construct a unigram language model by counting tokens in $\hat{D}'$ and use this language model as the simplest tokenizer.
We tokenize the validation and test dataset into the most plausible tokenization with the Viterbi algorithm.
Because the notation conflicts with Unigram explained in \S \ref{sec:downstream_model}, we denote the newly trained tokenizer as $\text{Unigram}^\text{OPT}$.

\paragraph{BI-Tag} As a neural tokenizer without vocabulary restriction, we employed a BiLSTM-CRF-based sequential tagging architecture~\cite{young2018recent}.
We convert the tokenization in $\hat{D}'$ into the BI-tagging format and train this architecture with it in a maximum of 200 epochs.
The size of character embeddings was 128 and the size of BiLSTM hidden vectors was 256.
We utilized an existing implementation\footnote{\url{https://github.com/jidasheng/bi-lstm-crf}}.

\paragraph{OpTok} As a conventional method for optimizing tokenization, we employed OpTok~\cite{hiraoka2021joint}. 
Because this method optimizes the unigram-based tokenizer by incorporating it into the downstream model, the tokenizer is not directly trained on $\hat{D}'$.
We used the existing imprimentation\footnote{\url{https://github.com/tatHi/optok4at}} and mostly followed the experimental settings explained in Section 4 of \newcite{hiraoka2021joint}.
For OpTok, we set $N=3$ due to its huge computation cost.

\section{Experimental Result}
\label{sec:performance}

Table \ref{tbl:downstream_result} shows the experimental results on test splits of each dataset\footnote{
In this paper, we report the averaged F1 scores obtained by three trials for BI-Tag, OpTok, and Proposed.
We reported the score of only one trial for $\text{Unigram}^\text{OPT}$ because the tokenizer construction of $\text{Unigram}^\text{OPT}$ is deterministic.
}.
The column named "\textbf{Original}" lists the performance of original downstream models $\theta$.
The proposed method and baselines optimized the tokenization using this original downstream model as post-processing.
The final performance obtained by each method is shown in columns named "\textbf{Optimized Tokenization}".

The result demonstrated that the proposed method and OpTok improve the performance against the original performance in all datasets.
Besides, the proposed method outperforms OpTok in four out of six comparable datasets\footnote{Since OpTok is applicable only to the unigram-based tokenizer, we could not evaluate it on BPE and MaxMatch.}.
We assume this because the proposed method can use more various types of tokenization candidates.
In fact, the proposed method can see $N=100$ candidates to create $\hat{D}'$ while OpTok is limited to using small numbers of tokenization candidates such as $N=3$ due to its huge computation cost.

The experimental results show that the simplest methods, Unigram-based models ($\text{Unigram}^\text{OPT}$), cannot obtain performance improvement in most datasets.
This result indicates that a more representative tokenizer is required for the proposed training strategy.
However, the the experimental results also show that the popular representative tokenizer (BI-Tag) did not improves the performance.
From the performance gap between BI-Tag and the proposed method, we conclude that the neural tokenizer with vocabulary restriction performs well to optimize tokenization in the post-processing environment.
We can use various representative tokenizers to learn $\hat{D}'$, and we expect that performance improvement can be obtained if they have a restriction not to output unknown tokens for the downstream model.

The rightmost column labeled as "\textbf{Oracle}" listed the scores obtained by extracting the best-performing tokenizations on the test splits.
As well as the dataset construction of $\hat{D}'$, we fed $N=100$ tokenization candidates of test inputs to the downstream model and extracted the best candidates using the test labels.
In other words, the scores in Oracle are upper bounds of this task (i.e., the task of optimizing tokenization as post-processing).
Of course, the test labels are unseen in the practical setting, and reaching these scores is not easy.
However, the performance gaps between the proposed method and Oracle show that we have much room to improve the performance by optimizing tokenization as the post-processing.

%% file: tables/datasets.tex
\begin{table}[t]
\centering
\begin{tabular}{p{0.15cm}p{1.15cm}|rrrr}
\hline
    & Dataset & Train   & Valid  & Test & \#L \\\hline
Zh  & Weibo   & 536,841 & 67,105 & 67,106 & 2 \\
    & Genre   & 312,000 & 39,000 & 39,000 & 13 \\
    & Rate    & 312,000 & 39,000 & 39,000 & 5 \\ \hdashline
Ja  & Twitter & 129,747 & 16,218 & 16,219 & 3 \\
    & WRIME   & 30,000 & 3,500 & 3,500 & 5 \\ \hdashline
En  & Twitter & 80,000 & 10,000 & 10,000 & 2 \\ \hline
\end{tabular}
\caption{
    Statistics of datasets showing the number of samples in each split and label variations (\#L).
    \label{tbl:dataset}
}
\end{table}


%% file: tables/downstream_result.tex
\begin{table*}[t]
\centering
\begin{tabular}{lll|r|rrrrr}
\hline
\multirow{2}{*}{Language} & \multirow{2}{*}{Dataset} & \multirow{2}{*}{Tokenizer} & \multicolumn{1}{l|}{} & \multicolumn{5}{c}{Optimized Tokenization}                                                                  \\
                          &                          &                            & Original     & $\text{Unigram}^\text{OPT}$        & BI-Tag         & OpTok                & \multicolumn{1}{r|}{Proposed}             & Oracle \\ \hline
Zh                   & Weibo                    & Unigram                    & 93.00                 & 92.99          & 92.83          & \u{{\textbf{93.15}}} & \multicolumn{1}{r|}{\textbf{93.13}}       & 97.76  \\
                          &                          & BPE                        & 92.72                 & 92.69          & 92.46          & -                    & \multicolumn{1}{r|}{\u{ \textbf{92.84}}} & 98.18  \\ 
                          &                          & MaxMatch                   & 92.46                 & 92.60          & 92.24          & -                    & \multicolumn{1}{r|}{\u{ \textbf{92.69}}} & 98.03  \\\hdashline
                          & Genre                    & Unigram                    & 48.15                 & 48.13          & 47.65          & \textbf{48.18}       & \multicolumn{1}{r|}{\u{ \textbf{48.24}}} & 71.94  \\
                          &                          & BPE                        & 48.23                 & 47.63          & 46.83          & -                    & \multicolumn{1}{r|}{\u{ \textbf{48.12}}} & 76.02  \\
                          &                          & MaxMatch                   & 48.16                 & 47.46          & 46.69          & -                    & \multicolumn{1}{r|}{\u{ \textbf{47.86}}} & 74.51  \\\hdashline
                          & Rate                     & Unigram                    & 47.92                 & \textbf{47.97} & \textbf{48.58} & \textbf{47.96}       & \multicolumn{1}{r|}{\u{ \textbf{48.75}}} & 79.85  \\
                          &                          & BPE                        & 47.89                 & 47.81          & 47.68          & -                    & \multicolumn{1}{r|}{\u{ \textbf{48.89}}} & 84.43  \\
                          &                          & MaxMatch                   & 47.82                 & \textbf{47.99} & \textbf{48.21} & -                    & \multicolumn{1}{r|}{\u{ \textbf{49.15}}} & 75.53  \\ \hline
Ja                  & Twitter                  & Unigram                    & 86.28                 & \textbf{86.30} & 83.21          & \textbf{86.29}       & \multicolumn{1}{r|}{\u{ \textbf{86.36}}} & 94.57  \\
                          &                          & BPE                        & 85.97                 & 85.91          & 83.79          & -                    & \multicolumn{1}{r|}{\u{ \textbf{86.02}}} & 95.45  \\
                          &                          & MaxMatch                   & 86.06                 & 85.99          & 84.78          & -                    & \multicolumn{1}{r|}{\u{ \textbf{86.08}}} & 95.45  \\\hdashline
                          & WRIME                    & Unigram                    & 44.83                 & 44.76          & 43.35          & \textbf{45.00}       & \multicolumn{1}{r|}{\u{ \textbf{45.41}}} & 75.05  \\
                          &                          & BPE                        & 43.92                 & 43.90          & 43.20          & -                    & \multicolumn{1}{r|}{\u{ \textbf{44.73}}} & 80.56  \\
                          &                          & MaxMatch                   & 43.77                 & 43.74          & 43.19          & -                    & \multicolumn{1}{r|}{\u{ \textbf{44.13}}} & 75.87  \\ \hline
En                   & Twitter                  & Unigram                    & 77.50                 & 77.39          & 77.22          & \u{ \textbf{77.77}} & \multicolumn{1}{r|}{\textbf{77.64}}       & 90.57  \\
                          &                          & BPE                        & 76.21                 & \textbf{76.34} & \textbf{76.55} & -                    & \multicolumn{1}{r|}{\u{ \textbf{76.71}}} & 96.10  \\
                          &                          & MaxMatch                   & 76.15                 & 75.86          & 75.61          & -                    & \multicolumn{1}{r|}{\u{ \textbf{76.36}}} & 96.38  \\ \hline
\end{tabular}
\caption{
    Experimental results on each text classification dataset (F1\%, average of 3 runs, test-split).
    Scores that surpass the ones of original downstream models $\theta$ (without optimizing tokenization) are highlighted in bold.
    Besides, the highest scores among the baselines are highlighted with underline.
    \label{tbl:downstream_result}
}
\end{table*}

%% file: sections/6_discussion.tex
\section{Discussion}
This section analyzes the feature of the proposed method.
For the analysis, we focus on the setting where the downstream model is trained using Unigram tokenizers because it mostly performed the best in the experiment (\S \ref{sec:experiments}).

\subsection{Tokenization Reproducibility}
\label{sec:reproducibility}
\input{tables/tokenization_acc.tex}

In the proposed training strategy, we train additional tokenizers with $\hat{D}'$.
The appropriate tokenization depends on the context and the target label.
In other words, a token that improves the performance of one input is not always applicable to other inputs.
This means that the appropriate tokenization in $\hat{D}'$ is not consistent, and tokenizers with poor capability could not reproduce even the training tokenization data $\hat{D}'$.

Table \ref{tbl:tokenization_acc} shows the reproducibility of tokenization in training splits by three tokenizers: \textbf{$\text{Unigram}^\text{OPT}$}, \textbf{BI-Tag}, and \textbf{Proposed}\footnote{
OpTok is out of focus in this analysis because it does not use $\hat{D}'$ for the training.
}.
We evaluated the $\text{Unigram}^\text{OPT}$ tokenizer with a unigram language model built by counting the number of tokens in $\hat{D}'$.
As for BI-Tag and the proposed method, we evaluated their models trained with $\hat{D}'$ in 200 epochs.
We measured reproducibility as the accuracy of tokenization after converting tokenization into BI-tagging format.

The result shows that the neural-based tokenizer (i.e, BI-Tag and Proposed), not surprisingly, can reproduce the training tokenization data.
We consider that this difference in the capability of representation brought the performance improvement of the proposed method against $\text{Unigram}^\text{OPT}$ and OpTok based on the unigram-based tokenizer in \S \ref{sec:experiments}.
BI-tag scores the best in most datasets because the proposed method is less representative than BI-Tag due to vocabulary restriction.
This means that the performance decrease of BI-Tag in \S \ref{sec:experiments} is caused by other reasons examined in the following subsection.

\subsection{Unknown Token Ratio}
\label{sec:unk_rate}
\input{tables/tokenization_unk.tex}
We introduced the vocabulary-restricted neural tokenizer in \S \ref{sec:tokenizer_with_restriction}.
This is because we hypothesized that the final performance would be damaged if the trained tokenizer outputs many unknown tokens.
This subsection investigates the actual appearance ratio of unknown tokens by each tokenizer.

Table \ref{tbl:tokenization_unk} shows the ratio of unknown tokens in the outputs of three tokenizers (i.e., $\text{Unigram}^\text{OPT}$, BI-Tag, and Proposed) on validation splits.
We tokenized the text of a validation split with these tokenizers and counted the number of tokens that are not in the vocabulary of the downstream model $V_\theta$.

$\text{Unigram}^\text{OPT}$ and the proposed method theoretically cannot output tokenization that includes the unknown tokens except the case where the input contains unknown characters\footnote{
    Unknown character is not rare for some languages using ideogram (e.g., \textit{hanzi} in Chinese) because they have more than thousands of characters.
}.
Therefore, the appearance ratio of unknown tokens in validation splits by these two methods was almost zero.
In contrast, BI-Tag outputs a maximum of 11.5\% unknown tokens in the validation splits because it does not have any output restriction.
Such unknown tokens cannot be encoded into token embeddings, and we consider this higher ratio of unknown outputs degrades the final performance of BI-Tag in \S \ref{sec:experiments}.
In fact, the performance of BI-Tag in Table \ref{tbl:downstream_result} was lower than the one of Original.

From the results of \S \ref{sec:reproducibility} and \ref{sec:unk_rate}, we conclude that it is reasonable to use a neural tokenizer with vocabulary restriction in terms of representation capability and output controllability, especially in the task of optimizing tokenization as post-processing.

\subsection{Effect of $N$}
\label{sec:effect_of_n}

\begin{figure}[t]
\centering
\includegraphics[width=7.7cm]{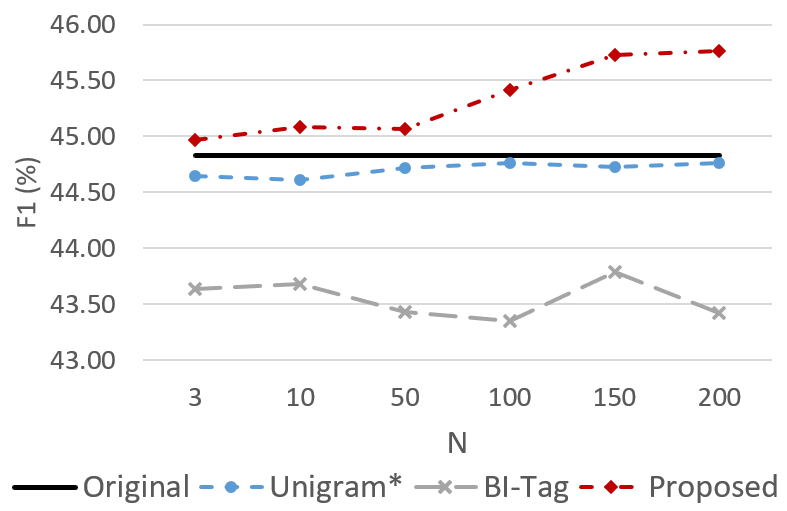}
\caption{
    Performance difference against $N$ on the test split of WRIME.
}
\label{fgr:diff_n}
\end{figure}

\input{tables/statistics}

The proposed method builds the additional training data $\hat{D}'$ by picking the best-performing tokenization from $N$ candidates (\S \ref{sec:creating_dataset}).
With a higher value of $N$, we can explore appropriate tokenization from the larger searching space of the original training data $D$.
This might be helpful to construct more effective $\hat{D}'$ to acquire better tokenizers.
This section investigates the effect of different $N = \{3, 10, 50, 100, 150, 200\}$ on the downstream performance with the WRIME dataset.
We utilized this dataset for the analysis because the final performance was the lowest among six datasets.
We used the model trained with the Unigram tokenizer as well as the other analysis in this section.

Figure \ref{fgr:diff_n} shows that the proposed method can slightly improve the performance even with $N=3$ against the original performance.
The actual value with $N=3$ was 44.97 which is comparable with the existing method, OpTok in Table \ref{tbl:downstream_result}.
This result indicates that the proposed method can be an good alternative to the existing method.
The result also shows that the performance of the proposed method is increased with higher $N$ up to $N=150$.
This result implies that the proposed method has a capability to improve performance with a training dataset created with a large number of tokenization candidates.

The scores of $\text{Unigram}^\text{OPT}$ are slightly increased with higher $N$ compared to lower $N$.
This implies that the $\hat{D}'$ has the potential to improve the final performance but $\text{Unigram}^\text{OPT}$ does not have capability enough to utilize it.
In contrast, the performance of BI-Tag is stably lower than the original performance possibly because it does not have vocabulary restriction.

\subsection{Tokenization Statistics}
This subsection reports a quantitative analysis of tokenization obtained by each baseline trained on $\hat{D}'$.
The left of Table \ref{tbl:stat} shows the averaged number of tokens in tokenized inputs.
The higher value indicates that the tokenized sequence includes more tokens.
In other words, the input sequence is tokenized into more fine-grained tokens.
In the training split, the newly created data $\hat{D}'$ contains longer sequences than the original data.
This is not surprising because the most plausible tokenization basically includes the longest tokens and we create $\hat{D}'$ from $N$ tokenization candidates.
As for the validation split, the result shows that the proposed method trained on $\hat{D}'$ yields a bit longer sequence than the original sequence.
The number of tokens outputted by $\text{Unigram}^\text{OPT}$ is almost the same as the one of the original sequence even though $\text{Unigram}^\text{OPT}$ is newly constructed with $\hat{D}'$.
In contrast, BI-Tag outputs the most fine-grained tokenization by yielding many B tags.

The right side of Table \ref{tbl:stat} reports perplexity measured on each tokenized data to investigate the peakiness of the distribution of token frequency.
In the training split, $\hat{D}'$ has higher perplexity than the one of the original (this is not surprising again for the same reason above).
This implies that the tokenization in $\hat{D}'$ is inconsistent and specific to each input.
In the validation split, similarly to the analysis of token length above, the perplexity of the proposed method tends to be lower than the one of the original.
The perplexity of $\text{Unigram}^\text{OPT}$ is almost the same as the original one, while the one of BI-Tag is remarkably higher than the original.

These two quantitative investigations suggest that $\text{Unigram}^\text{OPT}$ does not have the capability to strongly change the tendency of tokenization while BI-Tag changes too much.
Note that the capability of OpTok is theoretically the same as $\text{Unigram}^\text{OPT}$ because its tokenizer is limited to the unigram-based tokenizer.
On the other hand, the proposed method appropriately reflects the feature of $\hat{D}'$ to the validation split.
And we consider that this characteristic of the proposed method brought the performance improvement in Table \ref{tbl:downstream_result}.


%% file: tables/tokenization_acc.tex
\begin{table}[t]
\centering
\begin{tabular}{p{0.15cm}p{1.15cm}|rrr}
\hline
   & Dataset  & $\text{Unigram}^\text{OPT}$   & BI-Tag   & Proposed \\  \hline
Zh & Weibo   & 93.1    & 97.0 & 93.9   \\
       & Genre   & 89.4    & 93.6 & 90.9   \\
       & Rate    & 88.8    & 93.2 & 92.1   \\ \hdashline
Ja & Twitter & 93.8    & 97.2  & 98.3  \\
       & WRIME   & 93.2    & 100.0 & 99.9  \\ \hdashline
En   & Twitter & 96.7    & 97.0 & 97.5   \\ \hline
\end{tabular}
\caption{
    Reproducibiliy (accuracy) of each tokenization method on the training set.
    \label{tbl:tokenization_acc}
}
\end{table}

%% file: tables/tokenization_unk.tex
\begin{table}[t]
\centering
\begin{tabular}{p{0.15cm}p{1.15cm}|rrr}
\hline 
    & Dataset  & $\text{Unigram}^\text{OPT}$   & BI-Tag   & Proposed \\ \hline
Zh   & Weibo   & 0.0 & 1.0 & 0.0 \\
          & Genre   & 0.0 & 2.5 & 0.0 \\
          & Rate    & 0.0 & 2.3  & 0.0 \\ \hdashline
Ja  & Twitter & 0.0  &  4.2  & 0.0   \\
          & WRIME   & 0.1 & 11.5 & 0.1 \\ \hdashline
En   & Twitter & 0.0 & 5.6 & 0.0 \\ \hline
\end{tabular}
\caption{
    The appearance ratio of unknown tokens in the validation set.
    \label{tbl:tokenization_unk}
}
\end{table}

%% file: tables/statistics.tex
\begin{table*}[h!t]
\centering
\begin{tabular}{p{0.15cm}p{1.15cm}|rr:rrrr|rr:rrrr}
\hline
   &         & \multicolumn{6}{l|}{Averaged Number of Tokens}                             & \multicolumn{6}{l}{Perplexity}                          \\
   &         & \multicolumn{2}{l:}{Train} & \multicolumn{4}{l|}{Valid}  & \multicolumn{2}{l:}{Train} & \multicolumn{4}{l}{Valid}   \\
   &         & Org.        & $\hat{D}'$          & Org. & Uni.  & BI.   & Prp. & Org.       & $\hat{D}'$           & Org. & Uni.  & BI.   & Prp. \\ \hline
Zh & Weibo   & 22.0         & 23.8       & 22.1  & 22.1 & 30.7 & 22.4  & 119.9       & 164.9       & 20.9  & 19.9 & 23.6 & 22.1  \\
   & Genre   & 12.3         & 14.3       & 12.4  & 12.4 & 14.0 & 12.7  & 29.1        & 53.6        & 8.9   & 9.2  & 14.1 & 9.6   \\
   & Rate    & 12.3         & 14.5       & 12.4  & 12.4 & 14.2 & 13.0  & 29.1        & 55.9        & 8.9   & 9.3  & 15.5 & 10.4  \\ \hdashline
Ja & Twitter & 20.4         & 22.5       & 20.6  & 20.6 & 22.5 & 21.4  & 27.6        & 38.2        & 9.9   & 9.9  & 13.1 & 11.2  \\
   & WRIME   & 18.7         & 20.6       & 26.0  & 26.1 & 26.8 & 26.8  & 9.4         & 14.3        & 5.4   & 5.3  & 6.2  & 5.9   \\ \hdashline
En & Twitter & 21.9         & 23.6       & 22.3  & 22.3 & 22.4 & 22.5  & 51.5        & 57.5        & 21.8  & 22.0 & 22.6 & 21.7  \\ \hline
\end{tabular}
\caption{
    Averaged numbers of tokens and perplexities in tokenized datasets.
    Org., Uni., BI., and Prp. denote Original and Unigarm*, BI-Tag, and Proposed, respectively.
    \label{tbl:stat}
}
\end{table*}

%% file: sections/7_conclusion.tex
\section{Conclusion}
This paper have tackled the task of optimizing tokenization as post-processing.
In this task, we seek an appropriate tokenization for the already trained and fixed downstream models.
The proposed two-stepped training strategy\footnote{Our implementation will be available on GitHub soon.} enables us to use various tokenizers for optimization.
And the experimental results demonstrated that the proposed method with vocabulary restricted neural tokenizer stably outperformed the existing methods.
The experimental result also indicates the possibility of further performance improvement in this task (Proposed vs. Oracle in Table \ref{tbl:downstream_result}). 